\documentclass[sigconf,nonacm]{acmart}

\usepackage[utf8]{inputenc}
\usepackage[T1]{fontenc}
\usepackage{booktabs}
\usepackage{amsfonts}
\usepackage{amsmath}
\usepackage{microtype}
\usepackage[table]{xcolor}
\usepackage{graphicx}
\usepackage{adjustbox}
\usepackage{makecell}
\usepackage{multirow}
\usepackage{tabularx}
\usepackage{caption}
\usepackage{wrapfig}
\usepackage{enumitem}

\newcolumntype{C}[1]{>{\centering\arraybackslash}p{#1}}
\newcolumntype{L}[1]{>{\raggedright\arraybackslash}p{#1}}

\graphicspath{{figures/}}

\title{Anchor-Regularized Adaptation for Generalizable AI-Generated Image Detection with DINOv3}

\author{Hyeongjun Choi}
\affiliation{
  \institution{Sungkyunkwan University}
  \city{Suwon}
  \country{Republic of Korea}
}
\email{junhjun@g.skku.edu}

\author{Juhun Lee}
\affiliation{
  \institution{Sungkyunkwan University}
  \city{Suwon}
  \country{Republic of Korea}
}
\email{josejhlee@g.skku.edu}

\author{Davide Cozzolino}
\affiliation{
  \institution{University of Naples Federico II}
  \city{Naples}
  \country{Italy}
}
\email{davide.cozzolino@unina.it}

\author{Luisa Verdoliva}
\affiliation{
  \institution{University of Naples Federico II}
  \city{Naples}
  \country{Italy}
}
\email{verdoliv@unina.it}

\author{Simon S. Woo}
\affiliation{
  \institution{Sungkyunkwan University}
  \city{Suwon}
  \country{Republic of Korea}
}
\additionalaffiliation{
  \institution{Secure Machines Lab Inc}
  \city{Suwon}
  \country{Republic of Korea}
}
\email{swoo@g.skku.edu}

\renewcommand{\shortauthors}{Hyeongjun Choi, Juhun Lee, Davide Cozzolino, Luisa Verdoliva, and Simon S. Woo}

\begin{document}

\begin{abstract}

Recent works in AI-generated image detection have shown that careful training data alignment can improve generalization by removing spurious correlations. However, linear probes on frozen DINOv3 representations achieve remarkably strong performance even when trained on misaligned datasets. Motivated by this result, we analyze the underlying rationale and the limits of this generalization. We find that frozen DINOv3 performs well because its decisions rely on features that faithfully represent the space of authentic images. At the same time, its final layer is less effective at capturing the subtle pixel-artifact cues that can be emphasized by aligned training data. We further observe that naively mixing aligned and misaligned data during adaptation improves sensitivity to such cues but at the cost of distorting the pre-trained representation, limiting generalization.
To address this issue, we propose Anchor-Regularized Adaptation (ARA). We apply Low-Rank Adaptation to capture pixel-level artifacts while leveraging a frozen anchor classifier to avoid deviations from the original representation structure. This allows the model to exploit pixel-artifact cues without sacrificing generalization. Our method achieves state-of-the-art performance on nine diverse and challenging benchmarks, indicating that ARA enables complementary supervision from misaligned and aligned data for more effective detection.

\end{abstract}



\maketitle

\section{Introduction}
\label{sec:intro}

The rapid advancement of generative models has made it increasingly difficult to distinguish real images from synthetic fakes. This trend raises serious societal concerns, including misinformation and digital fraud~\cite{barrett2023identifying, epstein2023art,lin2024detecting,bontcheva2024generative}. Consequently, AI-generated image (AIGI) detection has emerged as a critical research problem. In response, the focus of AIGI detection has increasingly shifted beyond controlled benchmarks toward robustness on in-the-wild data~\cite{yan2025sanity,cavia2024real,guillaro2025bias,li2026artificial,cozzolino2024raising}.

\begin{figure*}[!t]
\centering
\includegraphics[width=0.98\linewidth]{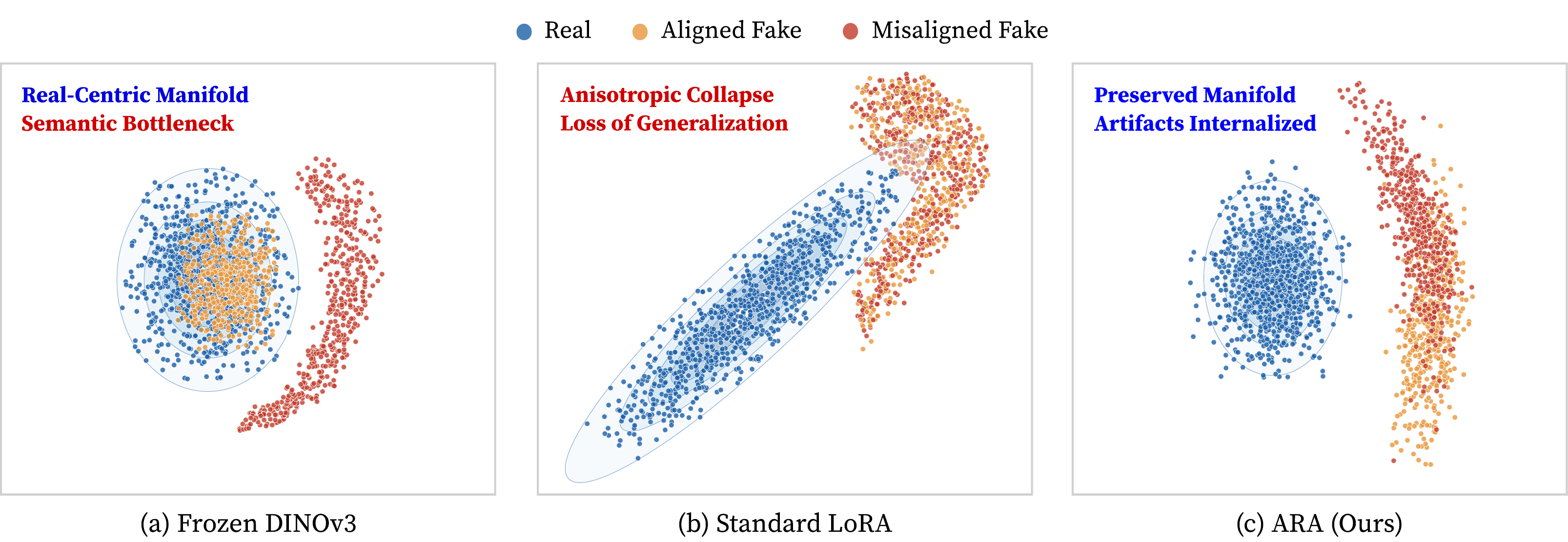}
\caption{Conceptual illustration of the structural trade-off in adapting DINOv3 for AIGI detection. (a) Frozen DINOv3 provides a strong authenticity
prior but struggles with fakes containing only subtle pixel-artifacts. (b) Naive fine-tuning can absorb such cues while distorting the pre-trained feature structure, degrading generalization. (c) ARA incorporates new artifact cues while better preserving the pre-trained structure.}
\Description{A three-panel overview contrasting frozen DINOv3, naive fine-tuning, and the proposed ARA method in terms of feature-space structure and forgery detection behavior.}
\label{fig:figure_1}
\vspace{-5pt}
\end{figure*}

Recent research in AIGI detection has advanced along two distinct trajectories. The first trajectory addresses the inherent vulnerabilities of training data. Conventional training datasets can contain statistical discrepancies between real and fake samples, such as content, format, or resolution, leading models to rely on spurious shortcuts rather than forensic traces~\cite{wang2020cnn,zhu2023genimage,grommelt2024fake}. This has led to a widespread effort toward \textit{dataset alignment}, primarily through VAE or diffusion-based reconstructions~\cite{chen2024drct,rajan2025aligned,guillaro2025bias,chen2025dual}. The second trajectory relies on the scale and capacity of modern backbones. While leveraging foundation models, particularly CLIP~\cite{radford2021learning}, is an established practice in the field~\cite{ojha2023towards,cozzolino2024raising,tan2025c2p,yan2024orthogonal,liu2024forgery}, the recent emergence of DINOv3~\cite{simeoni2025dinov3} marks a notable shift. This refocuses attention on inherent backbone capabilities by showing that simple linear probing over a frozen DINOv3 achieves state-of-the-art results on challenging benchmarks~\cite{zhou2026simplicity,huang2025rethinking}. In particular, Zhou et al.~\cite{zhou2026simplicity} achieve their breakthrough using GenImage~\cite{zhu2023genimage}, a dataset explicitly identified as misaligned in prior work~\cite{grommelt2024fake}. Frozen DINOv3 detectors trained on such misaligned data outperform several carefully aligned frameworks. This observation calls into question the prevailing belief that dataset misalignment is the primary bottleneck for real-world generalization. It therefore raises a critical question: does data alignment still matter in the era of powerful vision foundation models?

To answer this question, we systematically investigate the behavior of DINOv3-based detectors across misaligned and aligned data. Our analysis reveals two critical phenomena. First, we find that the remarkable generalization of frozen DINOv3 detectors is closely tied to a robust \textit{real-centric manifold}. Pre-trained on hundreds of millions of natural images, DINOv3 leverages a well-defined representation of real images. Even with only 1\% of the fake training samples retained, the model remains highly robust by identifying deviations from the well-defined real distribution. Second, we identify a structural limitation termed the \textit{semantic bottleneck}. When trained on strictly aligned data containing only subtle pixel-artifacts, a linear probe on the final layer features of frozen DINOv3 fails to fully capture these cues. Our layer-wise analysis shows that intermediate layers more effectively capture these low-level cues, but these cues are progressively suppressed during the semantic abstraction toward the final layer. This indicates that while DINOv3 possesses the representational capacity to detect subtle forgeries, its semantically oriented architecture reduces the salience of cues needed to detect low-level artifacts (Figure~\ref{fig:figure_1}(a)).

These findings reveal an architectural dilemma for universal detection across the forgery spectrum, from macroscopic semantic errors to microscopic pixel-artifacts. A naive attempt to overcome the semantic bottleneck involves \textit{decision fusion}, which averages predictions of intermediate and final layers. While this approach demonstrates that intermediate features hold complementary cues, simple decision-level integration does not fully incorporate these subtle artifacts into the model's core representation. Conversely, direct adaptation of the DINOv3 backbone through Low-Rank Adaptation (LoRA)~\cite{hu2022lora} allows the model to capture subtle artifacts. However, this often perturbs the pre-trained feature space, weakening the real-side structure that supports in-the-wild generalization (Figure~\ref{fig:figure_1}(b)). Thus, the challenge lies in capturing forgery artifacts without disrupting the feature geometry underlying robust generalization. To address this trade-off, we propose \textbf{Anchor-Regularized Adaptation (ARA)}. ARA introduces a frozen linear anchor classifier trained on the original DINOv3 representation to regularize backbone adaptation. During fine-tuning, this anchor encourages the adapted features to remain compatible with the real-side structure learned before adaptation, thereby limiting destructive representational drift. As a result, the model can encode subtle pixel-artifacts while preserving the feature geometry that underlies robust generalization, rather than increasing anisotropy in the feature space (Figure~\ref{fig:figure_1}(c)).
Overall, we make the following contributions:
\begin{itemize}[leftmargin=1.5em]
    \item We characterize the \textit{real-centric manifold} as an inherent strength of DINOv3-based detectors and the \textit{semantic bottleneck} as a critical structural weakness through a systematic analysis under misaligned and aligned data configurations.
    \item We identify the structural trade-off between capturing pixel-artifacts and preserving the feature space, and provide quantitative evidence that naive adaptation distorts the pre-trained geometry and limits generalization.
    \item We propose \textbf{ARA}, a regularization-based framework that achieves state-of-the-art performance across four standard benchmarks and five challenging in-the-wild datasets.
\end{itemize}
\section{Related Work}
\label{sec:related}

\begin{figure*}[!t]
\centering
\includegraphics[width=\linewidth]{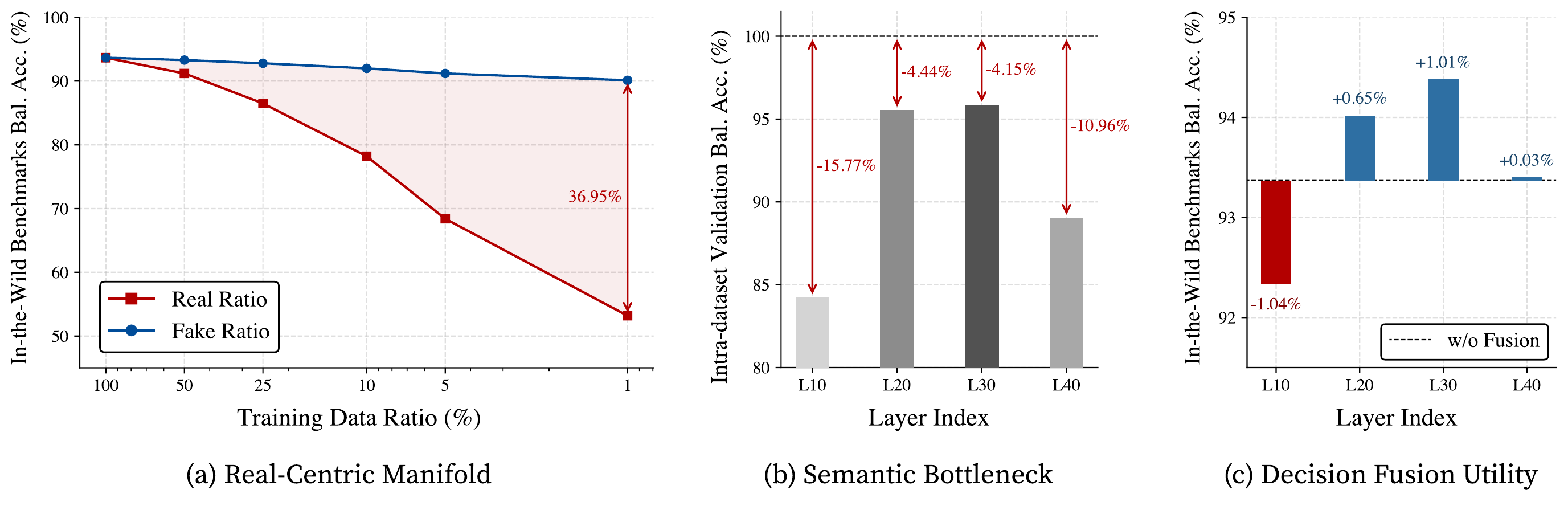}
\caption{Empirical analysis of frozen DINOv3. (a) Reducing the amount of real training data substantially degrades performance, confirming reliance on a \textit{real-centric manifold}. (b) Layer-wise probing on aligned data reveals a \textit{semantic bottleneck} at the final layer. (c) Fusing predictions from intermediate experts compensates for this bottleneck, improving in-the-wild generalization.}
\Description{Three plots showing data-ratio sensitivity, layer-wise probing accuracy, and decision-fusion gains for frozen DINOv3 under aligned and misaligned training regimes.}
\label{fig:figure_2}
\end{figure*}

\subsection{AIGI Detection and Dataset Alignment}
AI-generated image (AIGI) detection research has primarily focused on achieving robust generalization to unseen generators. Early studies rely on conventional benchmark datasets~\cite{wang2020cnn,zhu2023genimage} and focus on exploiting traces left by the generation process, such as low-level artifacts~\cite{cozzolino2019noiseprint, tan2024rethinking} or frequency inconsistencies~\cite{qian2020thinking, tan2024frequency}. However, Grommelt et al.~\cite{grommelt2024fake} expose a critical flaw in these conventionally used training datasets, revealing that misalignment between real and fake data forces models to learn spurious cues, ultimately causing generalization failures in the wild. This finding leads to a growing line of work on dataset alignment. To compel models to learn forensic artifacts rather than dataset biases, subsequent studies attempt to perfectly match resolution, semantic content, and compression characteristics. Representative works range from VAE reconstruction in Aligned~\cite{rajan2025aligned} to advanced pipelines in DRCT~\cite{chen2024drct}, B-Free~\cite{guillaro2025bias}, and DDA~\cite{chen2025dual}. Building on these efforts, recent works actively integrate aligned data to model real image distributions using them as hard samples~\cite{liu2025beyond}, enhance the perception capabilities of MLLMs~\cite{lin2025seeing}, or construct expert branches for pixel-artifacts~\cite{chen2025task}. Consequently, meticulously controlled aligned datasets now play a central role in the design of recent detection methodologies.

\subsection{Leveraging Vision Foundation Models}
In parallel with this trend, numerous studies rely on pre-trained Vision Foundation Models (VFMs). Ojha et al.~\cite{ojha2023towards} pioneer this approach by achieving high generalization performance through simple linear probing on CLIP~\cite{radford2021learning}. Subsequent research further leverages CLIP in AIGI detection~\cite{cozzolino2024raising} by adapting it to specific forgery cues~\cite{liu2024forgery, yan2024orthogonal}, injecting categorical concepts~\cite{tan2025c2p}, supplementing CNN-based models with high-level semantic cues~\cite{yan2025sanity}, or leveraging representations from intermediate encoder blocks~\cite{koutlis2024leveraging}. Furthermore, even dataset alignment studies~\cite{guillaro2025bias, chen2025dual} adopt DINO~\cite{oquab2023dinov2} as their backbone, highlighting the growing importance of VFMs in detector design. Within this context, the recent emergence of DINOv3~\cite{simeoni2025dinov3} has substantially advanced detection performance. Zhou et al.~\cite{zhou2026simplicity} demonstrate that simple linear probing on DINOv3 surpasses prior state-of-the-art techniques. Concurrently, Huang et al.~\cite{huang2025rethinking} show robust cross-generator generalization even with limited training samples. However, both studies report strong performance while relying on datasets previously identified as biased, such as GenImage~\cite{zhu2023genimage} and ForenSynth~\cite{wang2020cnn}. This outcome calls into question the narrative around dataset alignment. Consequently, these works primarily attribute detector performance to the backbone, emphasizing factors such as data exposure during pre-training~\cite{zhou2026simplicity} or a structural focus on low-level global features~\cite{huang2025rethinking}. However, these studies leave open how specific data alignment affects the representation space of DINOv3. To bridge this gap, our work shifts the focus from model architecture alone to the interaction between data alignment and DINOv3 representations.
\section{Preliminary Analysis}
\label{sec:analysis}

In this section, we conduct an in-depth analysis of DINOv3-based detectors to understand their behavior across two distinct data configurations: (1) a \textit{misaligned} environment using the standard GenImage~\cite{zhu2023genimage} benchmark, and (2) an \textit{aligned} environment where pixel-level cues are rigorously controlled. While aligned datasets from prior studies (e.g., DDA~\cite{chen2025dual}, B-Free~\cite{guillaro2025bias}) are valuable diagnostic tools, they often utilize real data from domains different from those in GenImage. This discrepancy potentially introduces domain shift as a confounding variable. To isolate the pure effect of dataset alignment on the feature space, we construct a domain-consistent \textit{AlignedGenImage} dataset. We use the real samples from GenImage (i.e., ImageNet~\cite{deng2009imagenet}) as the source for VAE-based reconstruction, ensuring a controlled comparison (see Section~\ref{subsubsec:alignedgenimage} for details).

\subsection{Real-Centric Manifold}
\label{subsec:real_manifold}
We adopt the DINOv3-7B linear probing detector trained on GenImage~\cite{zhou2026simplicity} as our strong baseline. To identify the underlying mechanism behind its robust generalization, we conduct an asymmetric data pruning experiment on the \textbf{GenImage} training set. Specifically, we independently scale down the volume of either real or fake images while keeping the opposite category fixed at its full scale. The results reveal a stark asymmetry in how the detector depends on real and fake training samples. As shown in Figure~\ref{fig:figure_2}(a), the average detection performance on in-the-wild datasets plummets from 93.37\% to 53.18\% when the proportion of real images is reduced to 1\%, indicating that the model loses a stable reference for authenticity. Conversely, the performance remains remarkably stable at 90.13\% even when the proportion of fake images is reduced to the same 1\%. This empirical evidence suggests that real images do not merely represent one side of a binary classification. Rather, they anchor the model's notion of natural image structure. Because DINOv3 is pre-trained on hundreds of millions of natural images, even minimal exposure to fake images (1\%) appears sufficient for the model to distinguish departures from this pre-trained notion of realness. We term this phenomenon the \textit{real-centric manifold}, and view it as a key factor behind the baseline's robust generalization.

\subsection{Semantic Bottleneck and Pixel-Artifacts}
\label{subsec:semantic_bottleneck}
Despite its generalization power, the baseline model exhibits a clear limitation in aligned settings, where detection performance significantly degrades for fake images containing only subtle pixel-artifacts~\cite{zhou2026simplicity}. To examine the degree of this limitation, we analyze whether the feature space of frozen DINOv3 can meaningfully capture fine-grained pixel-artifacts. Using our \textbf{AlignedGenImage} dataset, let $z_L = \Phi_L(x)$ denote the feature vector extracted from the $L$-th layer of the frozen backbone $\Phi$ for an input image $x$. To evaluate the accessibility of pixel-artifact cues across layers, we independently optimize a linear head $h_L$ on these features $z_L$ for $L \in \{10, 20, 30, 40\}$. As shown in Figure~\ref{fig:figure_2}(b), pixel-artifact information is strongest in the intermediate layers and markedly weaker at the final layer. Specifically, the convergence performance of linear heads follows the order: $h_{30}$ (95.85\%) $>$ $h_{20}$ (95.56\%) $>$ $h_{40}$ (89.04\%) $>$ $h_{10}$ (84.23\%). This pattern indicates that while intermediate processing captures subtle pixel-artifacts, their salience is progressively reduced during deeper semantic abstraction. We refer to this structural limitation as the \textit{semantic bottleneck} of DINOv3.

Building on this observation, we examine whether these pixel-artifacts can provide complementary cues for the baseline's blind spots. For each layer $L \in \{10, 20, 30, 40\}$, we treat the linear head optimized on AlignedGenImage as an expert head, $h_{L}$. We perform \textit{decision fusion} by averaging the probabilities from the baseline head $h_{\text{baseline}}$ (trained on standard GenImage) and the respective expert head $h_{L}$ as follows:
\begin{equation}
P_{\text{fusion}} = \frac{1}{2} \left( \sigma(h_{\text{baseline}}(z_{40})) + \sigma(h_{L}(z_{L})) \right),
\end{equation}
where $\sigma$ denotes the softmax function. As shown in Figure~\ref{fig:figure_2}(c), explicitly integrating predictions from heads that capture subtle pixel-level cues improves in-the-wild detection, peaking at 94.38\% (up from 93.37\%) when utilizing $h_{30}$. Notably, the improvement tracks the convergence of the expert head, following the order: $h_{30}$ (+1.01\%) $>$ $h_{20}$ (+0.65\%) $>$ $h_{40}$ (+0.03\%) $>$ $h_{10}$ (-1.04\%). This consistent pattern, also observed on DDA and B-Free, supports the view that fine-grained pixel-artifacts provide complementary cues for robust in-the-wild detection (see Supplementary Material~\ref{sec:cross_dataset_validation}).

\begin{figure}[t]
\centering
\includegraphics[width=0.85\linewidth]{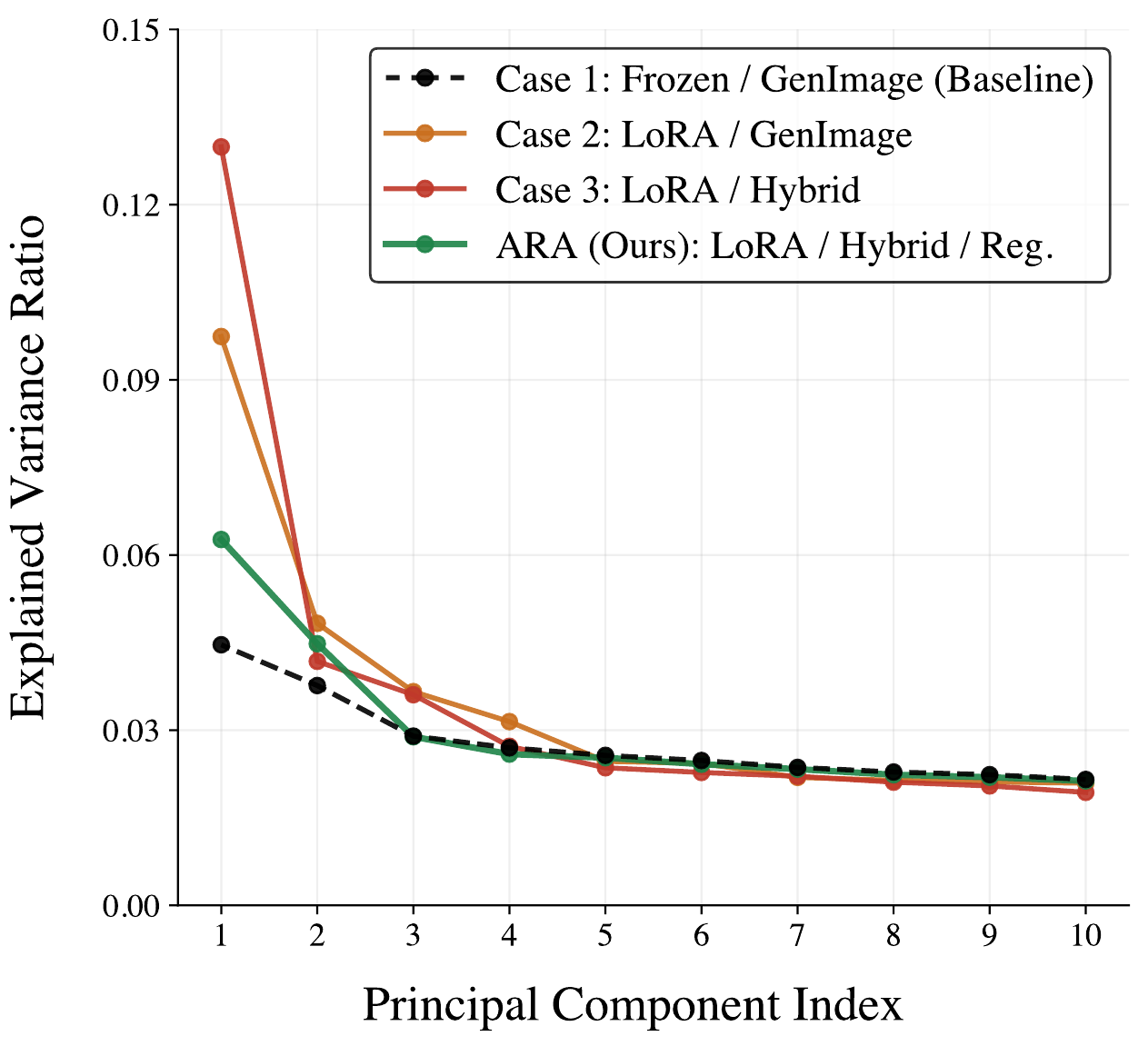}
\caption{Explained variance ratios of the top 10 principal components computed from real features. The increasing $PC_1/PC_2$ ratio in Cases 2 and 3 indicates stronger anisotropic distortion under backbone adaptation.}
\Description{Explained variance ratio of the top principal components across the compared adaptation regimes.}
\label{fig:figure_3}
\end{figure}
\section{Proposed Method}
\label{sec:method}

\subsection{Motivation}
\label{subsec:motivation}
The findings in Sections~\ref{subsec:real_manifold} and~\ref{subsec:semantic_bottleneck} suggest a critical dilemma in detector design. While frozen DINOv3 provides a robust \textit{real-centric manifold}, its \textit{semantic bottleneck} limits the detection of subtle forgery traces. To overcome this limitation, the model should better encode fine-grained pixel-artifacts in the final layer. However, because the frozen backbone lacks the capacity to exploit such cues at the final layer, doing so requires backbone adaptation. We therefore introduce LoRA~\cite{hu2022lora} and train it on a \textbf{hybrid} composition that pairs the original real images with a mixture of 50\% standard GenImage fakes and 50\% AlignedGenImage fakes. This setup allows the model to incorporate the hybrid forgery spectrum into a shared latent space. The key question is therefore whether this adaptation process preserves the geometric structure of real embeddings identified in Section~\ref{subsec:real_manifold}. In our setting, a more isotropic distribution of real features is preferable because it preserves variation across multiple directions rather than concentrating it along a narrow set of dominant directions~\cite{balestriero2025lejepa}. However, if adaptation is driven too strongly toward separating real images from the full hybrid forgery spectrum, it can overwrite the pre-trained real-side structure that supports robust discrimination between real and original fake images. As the feature space becomes increasingly specialized for this new objective, the real distribution may lose its original geometry and become more anisotropically distorted, weakening generalization. To examine this, we compare the representation space across three configurations: (Case 1) the frozen baseline, (Case 2) LoRA fine-tuning on GenImage, and (Case 3) LoRA fine-tuning on hybrid data. We characterize the structural changes using Principal Component Analysis (PCA) alongside three geometric metrics. We define a set of categories $\mathcal{C} = \{\text{real}, \text{original fake}, \text{aligned fake}\}$, where each category $c \in \mathcal{C}$ corresponds to a feature subset $\mathcal{Z}_c$. Let $\mu_c^{(k)}$, $D_c^{(k)}$, and $\Sigma_c^{(k)}$ denote the centroid, intra-class dispersion, and covariance matrix of category $c$ in Case $k$, respectively. For two categories $c_i$ and $c_j$, we further define the pooled covariance as $\tilde{\Sigma}_{c_i,c_j}^{(k)} = \Sigma_{c_i}^{(k)} + \Sigma_{c_j}^{(k)}$. Our geometric metrics are defined as follows:

\begin{table}[t]
\centering
\caption{Quantitative analysis of feature space geometry.}
\begin{adjustbox}{width=\columnwidth}
\begin{tabular}{l|cccccc}
\toprule
\multirow{2}{*}{\textbf{Method}} & \multicolumn{3}{c}{\textbf{Manifold Integrity} ($\downarrow$)} & \multicolumn{3}{c}{\textbf{Fisher Ratio} ($\uparrow$)} \\
\cmidrule(lr){2-4} \cmidrule(lr){5-7}
& $PC_1/PC_2$ & Drift & Dispersion & $F_{\text{real}, \text{original fake}}$ & $F_{\text{real}, \text{aligned fake}}$ & $F_{\text{original fake}, \text{aligned fake}}$ \\
\midrule
Case 1& 1.19 & 0.00 & 8.85 & 21.36 & 5.53 & 17.03 \\
Case 2& 2.02 & \textcolor{red}{5.24} & 8.74 & 107.71 & 5.98 & 15.21 \\
Case 3& \textcolor{red}{3.11} & 2.83 & \textcolor{red}{9.60} & 43.88 & 9.79 & 11.38 \\
\midrule
\textbf{ARA (Ours)} & \textbf{1.40} & \textbf{0.38} & \textbf{8.66} & 49.41 & 8.06 & 12.41 \\
\bottomrule
\end{tabular}
\end{adjustbox}
\label{tab:table_1}
\end{table}
\begin{table}[t]
\centering
\caption{Comparison of balanced accuracy (\%) on diverse benchmarks.}
\begin{adjustbox}{width=\columnwidth}
\small
\setlength{\tabcolsep}{3pt}
\begin{tabular}{l|C{10mm}C{10mm}C{10mm}C{10mm}C{10mm}|C{10mm}}
\toprule
\makecell{\textbf{Method}\\}
& \makecell{\textbf{AIGC}\\\textbf{Detect}}
& \makecell{\textbf{AIGI-}\\\textbf{Holmes}}
& \makecell{\textbf{AIGI-}\\\textbf{Now}}
& \makecell{\textbf{Synth}\\\textbf{buster}}
& \makecell{\textbf{In-the-}\\\textbf{Wild}}
& \makecell{\textbf{Avg.}\\} \\
\midrule
Case 1& 93.4 & 97.0 & 85.8 & 94.9 & 93.4 & 92.9 \\
Case 2& 94.9 & 97.0 & \textcolor{red}{74.5} & 98.3 & \textcolor{red}{92.7} & \textcolor{red}{91.5} \\
Case 3& 96.3 & 96.8 & \textcolor{red}{80.7} & 96.2 & \textcolor{red}{92.7} & \textcolor{red}{92.5} \\
\midrule
\textbf{ARA (Ours)} & \textbf{98.5} & \textbf{98.2} & \textbf{87.1} & \textbf{98.5} & \textbf{96.9} & \textbf{95.8} \\
\bottomrule
\end{tabular}
\end{adjustbox}
\label{tab:table_2}
\end{table}
\begin{figure*}[t]
\centering
\includegraphics[width=\linewidth]{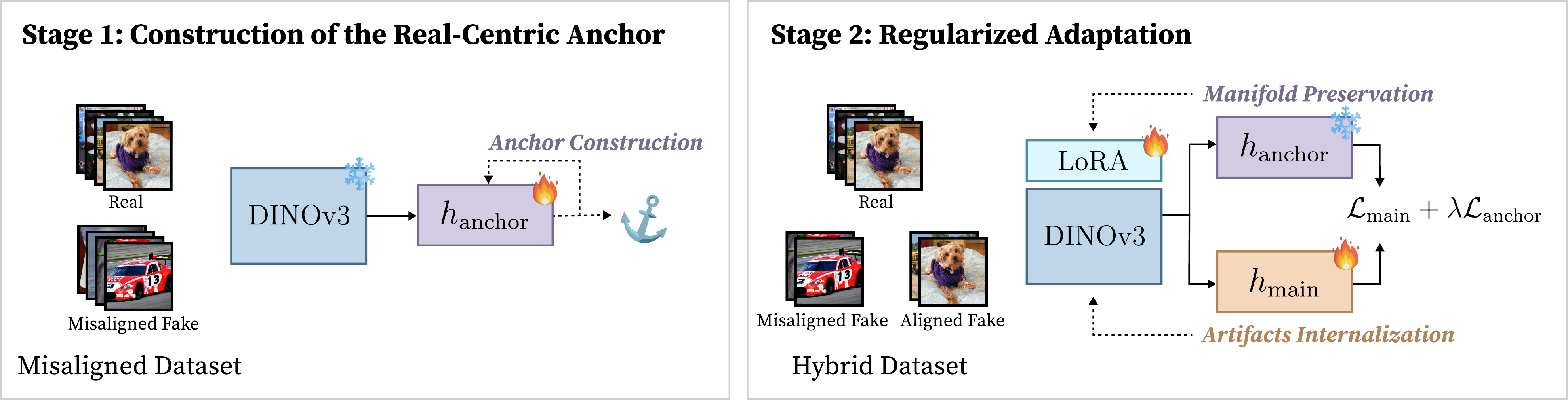}
\caption{Overview of the Anchor-Regularized Adaptation (ARA) framework. In Stage 1, a linear anchor head $h_{\text{anchor}}$ is trained on frozen DINOv3 final layer features using GenImage and then fixed as a geometric reference. In Stage 2, the DINOv3 backbone is adapted with LoRA on the hybrid dataset, where standard GenImage fakes and AlignedGenImage fakes are both treated as the binary \textit{fake} class. The resulting features are supervised by both a main classification loss $\mathcal{L}_{\text{main}}$ through $h_{\text{main}}$ and an anchor regularization loss $\mathcal{L}_{\text{anchor}}$ through the frozen head $h_{\text{anchor}}$.}
\Description{A pipeline diagram showing Stage 1 anchor construction on frozen DINOv3 features and Stage 2 LoRA adaptation regularized by the frozen anchor during hybrid-data training.}
\label{fig:figure_4}
\vspace{3pt}
\end{figure*}

\begin{itemize}[leftmargin=1.5em]
\item \textbf{Relative Drift} ($\Delta\mu_{\text{real}}^{(k)} = \| \mu_{\text{real}}^{(k)} - \mu_{\text{real}}^{(1)} \|_2$) measures the $\ell_2$ displacement of the real feature centroid relative to the frozen baseline (Case 1).

\item \textbf{Intra-Class Dispersion} ($D_{c}^{(k)} = \frac{1}{N_c} \sum_{z_i \in \mathcal{Z}_{c}} \| z_i - \mu_{c}^{(k)} \|_2$) quantifies the average $\ell_2$ distance of samples from their category centroid to assess the tightness of each distribution.

\item \textbf{Fisher Ratio} ($F_{c_i,c_j}^{(k)} = \sqrt{(\mu_{c_i}^{(k)} - \mu_{c_j}^{(k)})^\top (\tilde{\Sigma}_{c_i,c_j}^{(k)})^{-1} (\mu_{c_i}^{(k)} - \mu_{c_j}^{(k)})}$) measures class separation between two categories $c_i, c_j \in \mathcal{C}$ while accounting for their within-class covariance. We report the square root form for readability.
\end{itemize}
Additional details on the PCA protocol and geometric metric computation are provided in Supplementary Material~\ref{sec:geometric_details}.

In our subsequent analysis, we analyze the PCA eigenvalue distributions, the relative drift and dispersion of real features, and the Fisher ratios between category pairs. This allows us to evaluate how backbone adaptation alters the stability and cohesion of the real-side representation, as shown in Figure~\ref{fig:figure_3} and Table~\ref{tab:table_1}:

\vspace{1mm}
\noindent\textbf{Case 1: Geometric Baseline.}
Serving as our geometric baseline, Case 1 exhibits a relatively isotropic space, as indicated by a low $PC_1/PC_2$ ratio of 1.19. However, this structural integrity coincides with limited sensitivity to fine-grained pixel-artifacts, as reflected by the low Fisher ratio $F_{\text{real},\text{aligned fake}}^{(1)} = 5.53$.

\vspace{1mm}
\noindent\textbf{Case 2: Initial Adaptation.} 
The introduction of LoRA within the GenImage environment already modifies the representation structure. $PC_1$ increases by approximately 2.4$\times$, causing the $PC_1/PC_2$ ratio to rise to 2.02. This shift suggests that the manifold begins to stretch along a single dominant axis. Furthermore, a real drift of 5.24 indicates that even simple fine-tuning can substantially alter the original pre-trained coordinate system.

\vspace{1mm}
\noindent\textbf{Case 3: Anisotropic Collapse.} 
The most pronounced structural distortion appears under LoRA adaptation on the hybrid data. Although the model better captures pixel-artifacts, as reflected by the increased Fisher ratio $F_{\text{real},\text{aligned fake}}^{(3)} = 9.79$, this gain is accompanied by substantial geometric distortion. The $PC_1/PC_2$ ratio rises to 3.11, indicating that the representation is becoming increasingly concentrated along a narrow set of dominant directions. At the same time, the dispersion of real features increases to 9.60, suggesting reduced cohesion of the real-side structure. We refer to this combination of improved separation of aligned fakes and distorted real-side geometry as \textit{anisotropic collapse}.

These cases show that naive adaptation progressively distorts the real-side geometry, with drops in downstream performance across five benchmarks (Table~\ref{tab:table_2}). Case 2 already reduces the average accuracy from 92.9\% to 91.5\%, including a drop of 11.3 percentage points on AIGI-Now, and even hybrid training in Case 3 fails to recover the frozen baseline on the most challenging in-the-wild datasets. This pattern indicates that unconstrained backbone adaptation can over-specialize the representation toward training-specific artifacts and impair stability in out-of-distribution environments.

Across the analysis, three observations emerge: (1) the generalization of DINOv3-based detectors trained on misaligned data is strongly tied to a robust \textit{real-centric manifold}, (2) a \textit{semantic bottleneck} suppresses subtle pixel-artifact cues at the final layer, despite their potential to complement the model's blind spots, and (3) direct backbone adaptation can induce \textit{anisotropic collapse}, which is associated with weaker generalization. This highlights the core challenge of improving forensic sensitivity without destabilizing the geometry that supports generalization. Consequently, adaptation requires a geometric constraint that accounts for subtle forgery artifacts while limiting distortion of the real-centric manifold.

\begin{table*}[t]
\centering
\caption{Comparison of balanced accuracy (\%) on AIGCDetect.}
\begin{adjustbox}{max width=\textwidth}
\setlength{\tabcolsep}{1pt}
\small
\begin{tabular}{l|C{11mm}C{11mm}C{11mm}C{11mm}C{11mm}C{11mm}C{11mm}C{11mm}C{11mm}C{11mm}C{11mm}C{11mm}C{11mm}C{11mm}C{11mm}C{11mm}C{11mm}|C{11mm}}
\toprule
\textbf{Method} & 
\textbf{\makecell{ADM}} & 
\textbf{\makecell{DALLE2}} & 
\textbf{\makecell{GLIDE}} & 
\textbf{\makecell{Midjour.}} & 
\textbf{\makecell{VQDM}} & 
\textbf{\makecell{Big-\\GAN}} & 
\textbf{\makecell{Cycle-\\GAN}} & 
\textbf{\makecell{Gau-\\GAN}} & 
\textbf{\makecell{Pro-\\GAN}} & 
\textbf{\makecell{SDXL}} & 
\textbf{\makecell{SD1.4}} & 
\textbf{\makecell{SD1.5}} & 
\textbf{\makecell{Star-\\GAN}} & 
\textbf{\makecell{Style-\\GAN}} & 
\textbf{\makecell{Style-\\GAN2}} & 
\textbf{\makecell{WFR}} & 
\textbf{\makecell{Wukong}} & 
\textbf{\makecell{Avg.}} \\
\midrule
NPR~\cite{tan2024rethinking} & 43.8 & 20.0 & 41.2 & 53.4 & 48.4 & 53.1 & 76.6 & 42.2 & 58.7 & 59.6 & 55.1 & 55.0 & 67.4 & 57.9 & 54.6 & 58.8 & 57.4 & 53.1 \\
UnivFD~\cite{ojha2023towards} & 62.5 & 50.0 & 61.3 & 55.1 & 76.9 & 87.5 & 96.9 & 98.8 & 99.4 & 58.2 & 55.6 & 55.7 & 95.1 & 80.0 & 69.4 & 69.2 & 61.1 & 72.5 \\
FatFormer~\cite{liu2024forgery} & 80.2 & 68.5 & 91.1 & 54.4 & 88.0 & 99.2 & \textbf{99.5} & 99.1 & 98.5 & 71.7 & 67.5 & 67.2 & 99.4 & \textbf{98.0} & 98.8 & 88.3 & 75.6 & 85.0 \\
SAFE~\cite{li2025improving} & 49.5 & 49.5 & 53.0 & 49.0 & 50.2 & 52.2 & 51.9 & 50.0 & 50.0 & 49.8 & 49.7 & 49.8 & 50.1 & 50.0 & 50.0 & 49.8 & 50.3 & 50.3 \\
C2P-CLIP~\cite{tan2025c2p} & 71.6 & 52.3 & 73.5 & 56.6 & 73.7 & 98.4 & 96.8 & 98.8 & 99.3 & 62.3 & 77.5 & 76.9 & \textbf{99.6} & 93.1 & 79.4 & 94.8 & 79.4 & 81.4 \\
AIDE~\cite{yan2025sanity} & 52.9 & 51.1 & 60.2 & 49.8 & 69.3 & 70.1 & 93.6 & 60.6 & 89.0 & 49.6 & 51.6 & 51.0 & 72.1 & 66.5 & 59.0 & 80.6 & 54.5 & 63.6 \\
DRCT~\cite{chen2024drct} & 53.6 & 83.0 & 61.6 & 97.7 & 64.5 & 48.8 & 48.8 & 49.2 & 50.2 & 96.8 & 98.6 & 98.5 & 43.2 & 48.5 & 49.4 & 50.0 & 98.0 & 67.1 \\
B-Free~\cite{guillaro2025bias} & 76.4 & 74.9 & 70.8 & 94.3 & 89.2 & 91.5 & 66.7 & 96.1 & 95.4 & 99.6 & 99.4 & 99.3 & 81.0 & 75.4 & 71.0 & 58.9 & 99.4 & 84.7 \\
DDA~\cite{chen2025dual} & 90.7 & 93.9 & 88.8 & 93.9 & 67.8 & 81.1 & 64.8 & 81.6 & 73.1 & 98.0 & 96.7 & 96.7 & 67.9 & 67.8 & 76.7 & 49.8 & 96.8 & 81.5 \\
REM~\cite{liu2025beyond} & 98.1 & 99.3 & \textbf{98.6} & 96.5 & 99.3 & \textbf{99.4} & 98.1 & \textbf{99.8} & 97.4 & \textbf{99.8} & 99.3 & 99.5 & 90.2 & 95.1 & 96.3 & 98.4 & 99.5 & 97.9 \\
MIRROR~\cite{liu2026mirror} & 75.6 & 88.9 & 93.1 & 92.5 & 95.9 & 97.2 & 73.3 & 95.8 & 96.8 & \textbf{99.8} & \textbf{99.9} & \textbf{99.8} & 87.6 & 91.9 & 91.1 & 78.9 & \textbf{99.8} & 91.6 \\
\midrule
\textbf{ARA (Ours)} & \textbf{99.3} & \textbf{99.8} & 98.3 & \textbf{99.0} & \textbf{99.7} & \textbf{99.4} & 94.0 & 96.0 & \textbf{99.6} & \textbf{99.8} & 99.7 & 97.2 & 96.3 & 96.8 & \textbf{99.7} & \textbf{99.4} & 99.7 & \textbf{98.5} \\
\bottomrule
\end{tabular}
\end{adjustbox}
\label{tab:AIGCDetect}
\vspace{6pt}
\end{table*}

\subsection{Anchor-Regularized Adaptation}
To address this challenge, we propose \textbf{Anchor-Regularized Adaptation (ARA)}.
ARA is a simple yet effective learning framework that mitigates anisotropic collapse by using the real-centric manifold as an explicit regularization target during adaptation. Specifically, it constrains backbone fine-tuning with a frozen anchor classifier derived from the original DINOv3 representation. An overview of the framework is illustrated in Figure~\ref{fig:figure_4}.

\vspace{1mm}
\noindent\textbf{Stage 1: Construction of the Real-Centric Anchor.}
\label{subsec:stage1}
The first stage aims to construct a frozen anchor classifier grounded in the real-centric manifold of DINOv3 features. To this end, we use the misaligned GenImage dataset to optimize a linear head $h_{\text{anchor}}$ on top of the final layer features of DINOv3. As established in Section~\ref{subsec:real_manifold}, this classifier is trained on a representation that already exhibits a strong real-centric structure, rather than relying solely on dataset-specific forgery patterns. Once optimized, the parameters of $h_{\text{anchor}}$ remain fixed, allowing the frozen classifier to serve as a \textit{geometric anchor} during the subsequent fine-tuning stage. This anchor constrains representational drift away from the real-centric manifold established by the frozen model.

\vspace{1mm}
\noindent\textbf{Stage 2: Regularized Adaptation.}
\label{subsec:stage2}
The second stage involves joint optimization aimed at capturing pixel-artifacts while preserving the real-centric manifold. Training occurs on the hybrid dataset designed in Section~\ref{subsec:motivation}, where standard GenImage fakes and AlignedGenImage fakes share the same binary \textit{fake} label. To account for pixel-artifacts, we introduce LoRA to the DINOv3 backbone. For an input image $x$, let $z = \Phi_{\text{LoRA}}(x)$ denote the feature vector extracted from the final layer. This vector is passed to two independent classification heads to compute the following losses:
\begin{enumerate}[leftmargin=1.5em]
\item \textbf{Main Objective ($\mathcal{L}_{\text{main}}$):} The feature vector $z$ passes through a newly initialized main linear head $h_{\text{main}}$, where the cross-entropy loss $\mathcal{L}_{\text{main}}$ drives the model to capture the wide spectrum of forgery artifacts in the hybrid dataset.

\vspace{1mm}
\item \textbf{Geometric Regularizer ($\mathcal{L}_{\text{anchor}}$):} Simultaneously, $z$ passes through the frozen anchor $h_{\text{anchor}}$. The resulting cross-entropy loss $\mathcal{L}_{\text{anchor}}$ regularizes adaptation toward the structure defined by the frozen anchor over real and original fake samples. In doing so, it helps preserve the feature geometry that supports generalization, while encouraging aligned fake samples to be incorporated into the existing fake distribution.
\end{enumerate}

\noindent
The final loss function is defined as:
\begin{equation}
\mathcal{L}_{\text{total}} = \mathcal{L}_{\text{main}} + \lambda \mathcal{L}_{\text{anchor}},
\label{eq:ara_loss}
\end{equation}
where the hyperparameter $\lambda$ balances forgery sensitivity and manifold preservation. During inference, we leverage the fine-tuned backbone and the main head $h_{\text{main}}$ to predict image authenticity.

While naive LoRA fine-tuning on the hybrid data can capture new forgery cues, it also induces the anisotropic collapse identified as Case 3 in Section~\ref{subsec:motivation}. In contrast, ARA counteracts this geometric distortion through the $\mathcal{L}_{\text{anchor}}$ term. As shown in Figure~\ref{fig:figure_3} and Table~\ref{tab:table_1}, ARA preserves a more stable feature geometry than Case 3 while still retaining sensitivity to subtle forgery cues. This geometric advantage is reflected in the substantially stronger in-the-wild results of ARA in Table~\ref{tab:table_2}.

\begin{table}[t]
\centering
\caption{Comparison of balanced accuracy (\%) on Synthbuster.}
\begin{adjustbox}{max width=\columnwidth}
\setlength{\tabcolsep}{1pt}
\small
\begin{tabular}{l|C{11mm}C{11mm}C{11mm}C{11mm}C{11mm}C{11mm}C{11mm}C{11mm}C{11mm}|C{11mm}}
\toprule
\textbf{Method} &
\textbf{DALLE2} &
\textbf{DALLE3} &
\textbf{Firefly} &
\textbf{GLIDE} &
\textbf{Midjour.} &
\textbf{SD1.3} & \textbf{SD1.4} &
\textbf{SD2} &
\textbf{SDXL} &
\textbf{Avg.} \\
\midrule
NPR~\cite{tan2024rethinking} & 51.1 & 49.3 & 46.5 & 48.5 & 52.8 & 51.4 & 51.8 & 46.0 & 52.8 & 50.0 \\
UnivFD~\cite{ojha2023towards} & 83.5 & 47.4 & 89.9 & 53.3 & 52.5 & 70.4 & 69.9 & 75.7 & 68.0 & 67.8 \\
FatFormer~\cite{liu2024forgery} & 59.4 & 39.5 & 60.3 & 72.7 & 44.4 & 53.7 & 54.0 & 52.3 & 69.1 & 56.2 \\
SAFE~\cite{li2025improving} & 58.0 & 9.9 & 10.3 & 52.2 & 56.7 & 59.4 & 59.1 & 53.0 & 59.5 & 46.5 \\
C2P-CLIP~\cite{tan2025c2p} & 55.6 & 63.2 & 59.5 & 86.7 & 52.9 & 75.2 & 76.7 & 69.2 & 77.7 & 68.5 \\
AIDE~\cite{yan2025sanity} & 34.9 & 33.7 & 24.8 & 65.0 & 57.5 & 74.1 & 73.7 & 53.2 & 68.4 & 53.9 \\
DRCT~\cite{chen2024drct} & 3.9 & 36.8 & 13.5 & 21.6 & 98.9 & 94.9 & 93.8 & \textbf{99.9} & 96.4 & 62.2 \\
B-Free~\cite{guillaro2025bias} & 89.9 & 93.7 & \textbf{99.2} & 45.8 & 98.8 & \textbf{100.0} & 99.8 & 99.5 & 99.9 & 91.8 \\
DDA~\cite{chen2025dual} & 80.7 & 92.5 & 95.3 & 84.3 & \textbf{100.0} & 99.5 & 99.5 & 99.8 & \textbf{100.0} & 94.6 \\
MIRROR~\cite{liu2026mirror} & 98.0 & \textbf{99.9} & 91.6 & 96.5 & 98.3 & 99.8 & \textbf{100.0} & 99.2 & 99.5 & 98.1 \\
\midrule
\textbf{ARA (Ours)} & \textbf{98.2} & 99.3 & 92.8 & \textbf{99.3} & 98.7 & 99.6 & 99.6 & 99.3 & 99.6 & \textbf{98.5} \\
\bottomrule
\end{tabular}
\end{adjustbox}
\label{tab:Synthbuster}
\end{table}
\section{Experiments}
\label{sec:exp}

\begin{table*}[t]
\centering
\caption{Comparison of balanced accuracy (\%) on AIGI-Holmes.}
\begin{adjustbox}{width=\textwidth}
\small
\setlength{\tabcolsep}{2pt}
\begin{tabular}{l|C{14mm}C{14mm}C{14mm}C{14mm}C{14mm}C{14mm}C{14mm}C{14mm}C{14mm}C{14mm}|C{14mm}}
\toprule
\textbf{Method}
& \makecell{\textbf{FLUX}\\}
& \makecell{\textbf{Infinity}\\}
& \makecell{\textbf{Janus}\\}
& \makecell{\textbf{Janus-}\\\textbf{Pro-1B}}
& \makecell{\textbf{Janus-}\\\textbf{Pro-7B}}
& \makecell{\textbf{LlamaGen}\\}
& \makecell{\textbf{PixArt-XL}\\}
& \makecell{\textbf{SD3.5-L}\\}
& \makecell{\textbf{Show-o}\\}
& \makecell{\textbf{VAR}\\}
& \makecell{\textbf{Avg.}\\} \\
\midrule
CNNSpot~\cite{wang2020cnn} & 62.6 & 58.9 & 50.8 & 53.1 & 52.2 & 60.4 & 66.8 & 56.8 & 60.8 & 50.4 & 57.3 \\
FreqNet~\cite{tan2024frequency} & 89.4 & 92.4 & 44.3 & 43.9 & 43.3 & 87.5 & 88.5 & 89.4 & 91.0 & 68.8 & 73.9 \\
Gram-Net~\cite{liu2020global} & 69.6 & 60.4 & 49.1 & 49.1 & 49.1 & 62.7 & 62.7 & 70.1 & 77.1 & 55.7 & 60.6 \\
NPR~\cite{tan2024rethinking} & 96.8 & 98.3 & 49.5 & 49.5 & 49.5 & 94.7 & 89.9 & 93.7 & 98.8 & 73.0 & 79.4 \\
LaDeDa~\cite{cavia2024real} & 68.2 & 66.4 & 49.8 & 49.7 & 49.7 & 81.0 & 64.1 & 69.5 & 75.5 & 72.4 & 64.6 \\
UnivFD~\cite{ojha2023towards} & 86.8 & 89.8 & 57.5 & 67.4 & 57.5 & 91.5 & 92.5 & 89.8 & 91.1 & 59.7 & 78.4 \\
SAFE~\cite{li2025improving} & 93.2 & 96.5 & 48.2 & 48.2 & 48.3 & 91.3 & 89.6 & 94.0 & 96.2 & 91.1 & 79.7 \\
Effort~\cite{yan2024orthogonal} & 79.4 & 80.4 & 48.3 & 65.0 & 56.9 & 79.8 & 80.4 & 76.5 & 79.3 & 76.4 & 72.2 \\
DDA~\cite{chen2025dual} & \textbf{97.2} & 98.9 & 98.5 & 99.3 & 98.7 & 99.3 & 99.4 & 97.0 & 94.8 & 80.4 & 96.4 \\
OMAT~\cite{zhou2025breaking} & 94.7 & 95.5 & 65.1 & 75.6 & 64.1 & 96.7 & 96.9 & 95.7 & 96.9 & 95.9 & 87.7 \\
AIDE~\cite{yan2025sanity} & 94.4 & 98.7 & 91.2 & 98.9 & 97.8 & 99.4 & 98.6 & \textbf{99.4} & 98.0 & 93.6 & 97.0 \\
\midrule
\textbf{ARA (Ours)} & 91.0 & \textbf{99.7} & \textbf{99.8} & \textbf{99.6} & \textbf{99.8} & \textbf{99.7} & \textbf{99.8} & 93.5 & \textbf{99.6} & \textbf{99.7} & \textbf{98.2} \\
\bottomrule
\end{tabular}
\end{adjustbox}
\label{tab:AIGI-Holmes}
\vspace{5pt}
\end{table*}
\begin{table*}[t]
\centering
\caption{Comparison of balanced accuracy (\%) on AIGI-Now.}
\begin{adjustbox}{max width=\textwidth}
\small
\setlength{\tabcolsep}{3pt}
\begin{tabular}{l|C{7mm}C{7mm}C{7mm}C{7mm}C{7mm}C{7mm}C{7mm}C{7mm}C{7mm}C{7mm}C{7mm}C{7mm}C{7mm}C{7mm}C{7mm}C{7mm}C{7mm}C{7mm}|C{7mm}}
\toprule
\multirow{2}{*}{\makecell[l]{\textbf{Method}\\}} &
\multicolumn{2}{c}{\textbf{FLUX Dev}} &
\multicolumn{2}{c}{\textbf{FLUX Krea}} &
\multicolumn{2}{c}{\textbf{FLUX Kontext}} &
\multicolumn{2}{c}{\textbf{FLUX Pro}} &
\multicolumn{2}{c}{\textbf{GPT-4o}} &
\multicolumn{2}{c}{\textbf{Jimeng}} &
\multicolumn{2}{c}{\textbf{Keling}} &
\multicolumn{2}{c}{\textbf{Minimax}} &
\multicolumn{2}{c|}{\textbf{Nano Banana}} &
\multirow{2}{*}{\textbf{Avg.}} \\
\cmidrule(lr){2-3} \cmidrule(lr){4-5} \cmidrule(lr){6-7} \cmidrule(lr){8-9} \cmidrule(lr){10-11} \cmidrule(lr){12-13} \cmidrule(lr){14-15} \cmidrule(lr){16-17} \cmidrule(lr){18-19}
& \textit{Pix} & \textit{Sem} & \textit{Pix} & \textit{Sem} & \textit{Pix} & \textit{Sem} & \textit{Pix} & \textit{Sem} & \textit{Pix} & \textit{Sem} & \textit{Pix} & \textit{Sem} & \textit{Pix} & \textit{Sem} & \textit{Pix} & \textit{Sem} & \textit{Pix} & \textit{Sem} \\
\midrule
CNNSpot~\cite{wang2020cnn} & 91.9 & 50.0 & 55.0 & 50.0 & 84.3 & 50.2 & 53.5 & 50.0 & \textbf{99.0} & 50.1 & 52.3 & 50.0 & 97.3 & 50.1 & 60.3 & 50.4 & 98.5 & 49.9 & 63.5 \\
FreqNet~\cite{tan2024frequency} & 87.5 & 46.8 & 69.7 & 44.3 & 76.9 & 47.9 & 49.2 & 50.1 & 92.3 & 51.0 & 45.9 & 53.0 & 92.2 & 54.3 & 82.4 & 51.1 & 90.7 & 48.7 & 63.0 \\
Gram-Net~\cite{liu2020global} & 93.3 & 52.8 & 66.7 & 52.2 & 86.4 & 55.5 & 60.8 & 56.9 & 76.3 & 50.8 & 50.8 & 50.3 & 95.5 & 55.4 & 71.7 & 52.1 & 90.5 & 52.8 & 65.0 \\
NPR~\cite{tan2024rethinking} & 94.4 & 50.0 & 50.8 & 50.0 & 78.5 & 50.2 & 50.2 & 50.2 & 96.6 & 50.0 & 49.7 & 50.1 & 95.7 & 50.0 & 54.8 & 50.0 & 93.0 & 50.0 & 61.9 \\
LaDeDa~\cite{cavia2024real} & 58.6 & 49.8 & 49.7 & 49.7 & 56.0 & 50.2 & 49.6 & 49.5 & 74.5 & 49.9 & 49.5 & 49.7 & 66.1 & 50.1 & 50.9 & 50.2 & 76.6 & 50.5 & 54.5 \\
UnivFD~\cite{ojha2023towards} & 54.2 & 57.9 & 51.4 & 53.8 & 49.2 & 54.5 & 51.6 & 54.4 & 52.9 & 53.2 & 47.5 & 50.4 & 63.1 & 53.9 & 50.1 & 53.9 & 50.1 & 51.6 & 53.0 \\
SAFE~\cite{li2025improving} & 90.3 & 49.0 & 53.2 & 49.2 & 83.1 & 49.4 & 52.1 & 48.6 & 97.7 & 48.8 & 50.9 & 48.6 & 96.0 & 48.7 & 59.0 & 49.1 & 96.1 & 48.4 & 62.1 \\
Effort~\cite{yan2024orthogonal} & 78.9 & 67.9 & \textbf{79.6} & 66.9 & 72.8 & 61.0 & 68.8 & 69.0 & 75.3 & 58.0 & 52.2 & 55.5 & 78.2 & 67.7 & 77.2 & 68.7 & 79.6 & 65.7 & 69.1 \\
DDA~\cite{chen2025dual} & 91.6 & 51.2 & 59.4 & 49.9 & 82.7 & 52.9 & 76.6 & 55.0 & 92.3 & 65.4 & \textbf{87.0} & 65.4 & 96.1 & 64.6 & \textbf{83.3} & 50.5 & 81.6 & 56.2 & 70.1 \\
OMAT~\cite{zhou2025breaking} & 91.1 & 47.5 & 64.9 & 46.9 & 84.7 & 50.7 & 59.1 & 51.5 & 74.4 & 45.2 & 49.1 & 46.5 & 93.6 & 52.6 & 69.9 & 46.7 & 89.1 & 46.8 & 61.7 \\
AIDE~\cite{yan2025sanity} & \textbf{99.1} & 59.0 & 50.4 & 56.9 & \textbf{97.9} & 80.6 & 60.1 & 53.8 & 74.7 & 51.8 & 63.9 & 51.4 & \textbf{98.2} & 55.4 & 51.4 & 54.1 & \textbf{98.9} & 51.8 & 67.2 \\
\midrule
\textbf{ARA (Ours)} & 91.3 & \textbf{95.8} & 78.0 & \textbf{84.5} & 77.1 & \textbf{84.9} & \textbf{84.6} & \textbf{94.7} & 83.1 & \textbf{97.0} & 85.4 & \textbf{95.8} & 88.8 & \textbf{94.6} & 68.4 & \textbf{82.7} & 87.0 & \textbf{94.5} & \textbf{87.1} \\
\bottomrule
\end{tabular}
\end{adjustbox}
\label{tab:AIGI-Now}
\end{table*}

\subsection{Experimental Setup}
\label{subsec:setup}

\noindent\textbf{Datasets.}
We evaluate performance on nine different datasets, comprising four standard benchmarks (AIGCDetect~\cite{zhong2023patchcraft}, Synthbuster~\cite{bammey2023synthbuster}, AIGI-Holmes~\cite{zhou2025aigi}, and AIGI-Now~\cite{chen2025task}) and five in-the-wild benchmarks (Chameleon~\cite{yan2025sanity}, WildRF~\cite{cavia2024real}, SynthWildX~\cite{cozzolino2024raising}, AIGIBench~\cite{li2026artificial}, and B-Free-Viral~\cite{guillaro2025bias}). Standard benchmarks cover a broad range of generative model families, including GANs, diffusion models, and autoregressive models, and are used to evaluate cross-generator generalization. In contrast, in-the-wild benchmarks assess robustness under unconstrained real-world conditions. They consist of internet-sourced images with often-unknown generation and post-processing histories, providing challenging out-of-distribution evaluations under diverse real-world degradations.

\vspace{1mm}
\noindent\textbf{Evaluation Protocol.}
We report balanced accuracy following the convention adopted in most prior AIGI detection studies~\cite{guillaro2025bias,chen2025dual,zhou2026simplicity}. For the in-the-wild benchmarks, AIGCDetect, and Synthbuster, we adopt the evaluation protocol of MIRROR~\cite{liu2026mirror}. Our comparison includes multiple general detectors (NPR~\cite{tan2024rethinking}, UnivFD~\cite{ojha2023towards}, FatFormer~\cite{liu2024forgery}, SAFE~\cite{li2025improving}, C2P-CLIP~\cite{tan2025c2p}, and AIDE~\cite{yan2025sanity}), dataset alignment approaches (DRCT~\cite{chen2024drct}, B-Free~\cite{guillaro2025bias}, and DDA~\cite{chen2025dual}), and recent methods such as REM~\cite{liu2025beyond} and MIRROR~\cite{liu2026mirror}. To avoid inconsistencies from unofficial reproductions, we rely on benchmark results measured with official checkpoints whenever such results are available in the literature. Specifically, we source general detector performances from REM~\cite{liu2025beyond}, supplementing the omitted B-Free-Viral results with data from DDA~\cite{chen2025dual}. We draw results for the alignment approaches directly from MIRROR~\cite{liu2026mirror}. Additionally, we evaluate performance on the recently published AIGI-Holmes and AIGI-Now benchmarks by following the evaluation protocol of our primary baseline~\cite{zhou2026simplicity}. For these datasets, we compare our method against a broader suite of detectors, including CNNSpot~\cite{wang2020cnn}, FreqNet~\cite{tan2024frequency}, Gram-Net~\cite{liu2020global}, NPR~\cite{tan2024rethinking}, UnivFD~\cite{ojha2023towards}, SAFE~\cite{li2025improving}, LaDeDa~\cite{cavia2024real}, Effort~\cite{yan2024orthogonal}, DDA~\cite{chen2025dual}, OMAT~\cite{zhou2025breaking}, and AIDE~\cite{yan2025sanity}.

\vspace{1mm}
\noindent\textbf{AlignedGenImage Dataset Construction.}
\label{subsubsec:alignedgenimage}
The standard GenImage~\cite{zhu2023genimage} exhibits spurious differences between real images (ImageNet~\cite{deng2009imagenet}) and fake images (Stable Diffusion 1.4~\cite{rombach2022high}), such as variations in resolution, file format, and compression~\cite{grommelt2024fake}. To encourage the model to focus more directly on pixel-artifacts, we construct a strictly controlled AlignedGenImage dataset. Since existing aligned datasets from DDA~\cite{chen2025dual} and B-Free~\cite{guillaro2025bias} use MSCOCO~\cite{lin2014microsoft} as their real source, generating a new ImageNet-based dataset is essential for domain-consistent analysis. Following the methodology of DDA, we construct aligned fake samples by passing ImageNet images through the f8-ft-MSE VAE~\cite{stabilityai2022sdvaeftmse} while preserving their original resolution under the required 8$\times$ spatial constraints. Furthermore, we eliminate shortcuts caused by compression rate discrepancies. We achieve this by matching the JPEG compression level of each reconstructed image to that of its corresponding ImageNet source image, using the metadata provided by~\cite{grommelt2024fake}.

\vspace{1mm}
\noindent\textbf{Dataset Configuration for Fair Analysis.}
While our primary datasets are GenImage and AlignedGenImage, we additionally consider prior aligned datasets (DDA and B-Free) in Section~\ref{sec:analysis} to identify common trends across different data configurations. To ensure fairness, we establish a baseline scale of 51K real images, matching B-Free, the smallest dataset considered in our analysis. We subsample all datasets to contain exactly 51K real images and sample the same number of fake images. For aligned datasets, we sample fake images in direct pairs with their real counterparts, whereas we apply random sampling for the misaligned GenImage dataset. Consequently, all experimental datasets maintain a balanced ratio of 51K real and 51K fake images. The hybrid dataset used to train our method contains 51K real GenImage samples and an equal number of fake samples, with the fake set composed of a 50/50 mixture of standard GenImage and AlignedGenImage.

\begin{table*}[t]
\centering
\caption{Comparison of balanced accuracy (\%) on in-the-wild benchmarks.}
\begin{adjustbox}{max width=\textwidth}
\small
\setlength{\tabcolsep}{3pt}
\begin{tabular}{l|C{16mm}|C{12mm}C{12mm}C{12mm}|C{12mm}C{12mm}C{12mm}|C{18mm}C{18mm}|C{16mm}|C{16mm}}
\toprule 
\multirow{2}{*}{\makecell[l]{\textbf{Method}\\}} &
\multirow{2}{*}{\makecell{\textbf{Chameleon}\\}} &
\multicolumn{3}{c|}{\textbf{SynthWildX}} &
\multicolumn{3}{c|}{\textbf{WildRF}} &
\multicolumn{2}{c|}{\textbf{AIGIBench}} &
\multirow{2}{*}{\makecell{\textbf{B-Free}\\\textbf{Viral}}} &
\multirow{2}{*}{\makecell{\textbf{Avg.}\\}} \\
\cmidrule(lr){3-5} \cmidrule(lr){6-8} \cmidrule(lr){9-10}
& & DALLE3 & Firefly & Midjour. & Facebook & Reddit & Twitter & SocialRF & CommunityAI & & \\
\midrule
NPR~\cite{tan2024rethinking} & 59.9 & 43.6 & 61.3 & 44.5 & 78.1 & 61.0 & 51.3 & 53.3 & 55.1 & 49.5 & 55.8 \\
UnivFD~\cite{ojha2023towards} & 50.7 & 45.4 & 65.3 & 46.2 & 49.1 & 60.2 & 56.5 & 54.6 & 51.4 & 49.0 & 52.8 \\
FatFormer~\cite{liu2024forgery} & 51.2 & 46.5 & 61.6 & 48.3 & 54.1 & 68.1 & 54.4 & 56.9 & 51.9 & 50.0 & 54.3 \\
SAFE~\cite{li2025improving} & 59.2 & 49.4 & 48.2 & 49.6 & 50.9 & 74.1 & 37.5 & 58.4 & 54.5 & 50.5 & 53.2 \\
C2P-CLIP~\cite{tan2025c2p} & 51.1 & 56.9 & 61.4 & 53.0 & 54.4 & 68.4 & 55.9 & 56.4 & 51.0 & 50.0 & 55.9 \\
AIDE~\cite{yan2025sanity} & 63.1 & 63.4 & 48.8 & 51.9 & 57.8 & 71.5 & 45.8 & 57.3 & 53.7 & 53.1 & 56.6 \\
DRCT~\cite{chen2024drct} & 79.8 & 85.9 & 58.9 & 90.5 & 90.3 & 66.8 & 79.6 & 71.3 & 84.6 & 77.1 & 78.5 \\
Aligned~\cite{rajan2025aligned} & 61.3 & 49.6 & 53.9 & 52.4 & 48.4 & 54.0 & 40.6 & 51.0 & 60.2 & 38.1 & 51.0 \\
B-Free~\cite{guillaro2025bias} & 78.3 & 96.1 & 92.3 & 95.3 & 95.6 & 85.5 & 96.7 & 84.9 & 79.7 & 87.1 & 89.2 \\
DDA~\cite{chen2025dual} & 83.5 & 91.1 & 84.7 & 91.6 & 85.3 & 82.5 & 89.3 & 79.9 & 88.9 & 81.2 & 85.8 \\
REM~\cite{liu2025beyond} & 91.3 & 96.5 & 92.8 & 96.9 & 97.2 & 98.1 & 98.2 & 92.2 & 90.4 & - & 94.8 \\
MIRROR~\cite{liu2026mirror} & 90.7 & 95.9 & 88.4 & 94.9 & 97.1 & 96.6 & 96.4 & 87.6 & 93.4 & 83.0 & 92.4 \\
\midrule
\textbf{ARA (Ours)} & \textbf{95.4} & \textbf{97.8} & \textbf{95.2} & \textbf{97.6} & \textbf{97.8} & \textbf{98.3} & \textbf{98.5} & \textbf{96.8} & \textbf{97.0} & \textbf{94.9} & \textbf{96.9} \\
\bottomrule
\end{tabular}
\end{adjustbox}
\label{tab:in-the-wild}
\end{table*}

\vspace{1mm}
\noindent\textbf{Implementation Details.}
We employ DINOv3-ViT-7B, the largest model in the DINOv3 family, as the backbone for all experiments. All training and inference are conducted on a single NVIDIA RTX PRO 6000 Blackwell GPU. The training process consists of two stages. In Stage 1, we attach a linear head to the frozen DINOv3 backbone and train it on the standard GenImage dataset to construct the real-centric anchor. In Stage 2, we fine-tune the DINOv3 backbone using LoRA with a rank of 8 on the hybrid dataset. Both stages employ the AdamW~\cite{loshchilov2017decoupled} optimizer with a batch size of 128, with learning rates of 1e-3 for Stage 1 and 1e-4 for Stage 2. We train Stage 1 for 5 epochs, which is sufficient for the anchor head to reach stable convergence. We train Stage 2 for 1 epoch, as LoRA adaptation converges rapidly in our setting and longer training does not yield meaningful gains. Following~\cite{zhou2026simplicity}, we resize and center-crop all input images to 224$\times$224 without any data augmentation.

\subsection{Experimental Results}
\label{subsec:results}

\noindent\textbf{Performance on Standard Benchmarks.}
As shown in Table~\ref{tab:AIGCDetect} and Table~\ref{tab:Synthbuster}, ARA achieves state-of-the-art performance on both benchmarks, with average accuracies of 98.5\% on AIGCDetect and 98.5\% on Synthbuster. We further evaluate ARA on the recently introduced AIGI-Holmes and AIGI-Now benchmarks, which cover the most recent generative models. As shown in Table~\ref{tab:AIGI-Holmes} and Table~\ref{tab:AIGI-Now}, ARA achieves 98.2\% on AIGI-Holmes. Most notably, ARA achieves 87.1\% average accuracy on AIGI-Now, a benchmark designed to reflect the latest and most difficult generative models. This outperforms the closest competitor, DDA, by a significant margin of 17.0 percentage points. These results show high robustness against recent diffusion and autoregressive models.

\vspace{1mm}
\noindent\textbf{Performance on Challenging Benchmarks.}
These benchmarks evaluate real-world robustness using images from the web that contain unpredictable and complex degradations. As shown in Table~\ref{tab:in-the-wild}, ARA consistently achieves the highest average accuracy of 96.9\% across all five in-the-wild datasets. Our method achieves better results than MIRROR (92.4\%) and achieves the highest reported average performance on these benchmarks. Notably, ARA achieves the highest accuracy on every individual sub-dataset, indicating that the gain transfers consistently across heterogeneous in-the-wild settings rather than being confined to a single source. This suggests that ARA improves sensitivity to aligned forgery cues without sacrificing robustness under real-world distribution shift, further supporting the effectiveness of geometric regularization.

\subsection{Ablation Study}
\label{subsec:ablation}

\noindent\textbf{Impact of the Aligned Fake Ratio.}
We first investigate the optimal composition of the hybrid training dataset by varying the ratio $r$ of AlignedGenImage fakes within the fake portion of the hybrid set ($\lambda=1.0$). Table~\ref{tab:ablation}(a) shows that a balanced ratio of 0.5 yields the highest performance at 96.9\%. Performance drops sharply to 91.5\% when no aligned fakes are used ($r=0.00$), confirming that standard GenImage alone is insufficient to exploit subtle pixel-artifacts, while raising the ratio to $r=0.75$ slightly reduces performance to 96.5\%. These results support a balanced hybrid composition as the best trade-off between artifact sensitivity and generalization.

\vspace{1mm}
\noindent\textbf{Efficacy of the Anchor Weight.}
We next analyze the effect of the geometric regularizer by varying the anchor weight $\lambda$ (Table~\ref{tab:ablation}(b)). Setting $\lambda=0$ (naive LoRA without geometric constraints) yields the lowest performance at 92.7\%, supporting the instability of unconstrained adaptation discussed in Section~\ref{subsec:motivation}. Introducing the regularizer substantially improves performance, peaking at 96.9\% at $\lambda=1.0$. Increasing $\lambda$ beyond 1.0 slightly reduces performance, implying that excessive regularization restricts the model's capacity to adapt to subtle pixel-level cues. These results show that both the hybrid data composition and the anchor regularizer are necessary to reach the best in-the-wild performance.

\vspace{1mm}
\noindent\textbf{Additional Analyses.}
We provide further analyses in the Supplementary Material to examine our findings beyond the main ablations. Section~\ref{sec:regularization_variants} studies alternative anchor loss configurations and regularized adaptation and preservation baselines. Section~\ref{sec:backbone_scale} investigates the effect of backbone scale across different DINOv3 variants, while Section~\ref{sec:robustness} evaluates robustness to post-processing.

\begin{table}[t]
\centering
\caption{Comparison of balanced accuracy (\%) under different ablation settings on the in-the-wild benchmark average.}
\label{tab:ablation}
\small
\begin{minipage}[t]{0.46\columnwidth}
\vspace{0pt}
\begin{tabular}{C{15mm}|C{15mm}}
\toprule
\multicolumn{2}{c}{\textbf{(a) Aligned Fake Ratio}} \\
\midrule
$r=0.00$ & 91.5 \\
$r=0.25$ & 94.4 \\
$r=0.50$ & \textbf{96.9} \\
$r=0.75$ & 96.5 \\
\bottomrule
\end{tabular}
\end{minipage}
\hfill
\begin{minipage}[t]{0.46\columnwidth}
\vspace{0pt}
\begin{tabular}{C{15mm}|C{15mm}}
\toprule
\multicolumn{2}{c}{\textbf{(b) Anchor Weight}} \\
\midrule
$\lambda=0.0$   & 92.7 \\
$\lambda=0.5$ & 96.4 \\
$\lambda=1.0$ & \textbf{96.9} \\
$\lambda=2.0$ & 96.4 \\
\bottomrule
\end{tabular}
\end{minipage}
\end{table}
\section{Conclusion}
\label{sec:conclusion}

This paper provides a systematic analysis of DINOv3 for AIGI detection. We show that direct fine-tuning aimed at overcoming DINOv3's semantic bottleneck can distort the real-centric manifold that supports its robust generalization. Driven by this observation, we develop Anchor-Regularized Adaptation (ARA), which combines hybrid data adaptation with an anchor regularizer to capture subtle forgery cues while limiting distortion of the original feature space. Our results suggest that DINOv3 can be further improved through the complementary use of misaligned and aligned training data and an appropriate adaptation strategy.
\begin{acks}

This work was supported by Institute of Information \& Communications Technology Planning \& Evaluation (IITP) grant funded by the Korea government (MSIT) (No.RS-2024-00441762, Global Advanced Cybersecurity Human Resources Development).

\end{acks}

\bibliographystyle{ACM-Reference-Format}
\bibliography{references}

\clearpage
\appendix
\section{Supplementary Material}
\label{sec:supplement}

\subsection{Cross-Dataset Validation}
\label{sec:cross_dataset_validation}

\begin{figure}[h]
\centering
\includegraphics[width=1.0\linewidth]{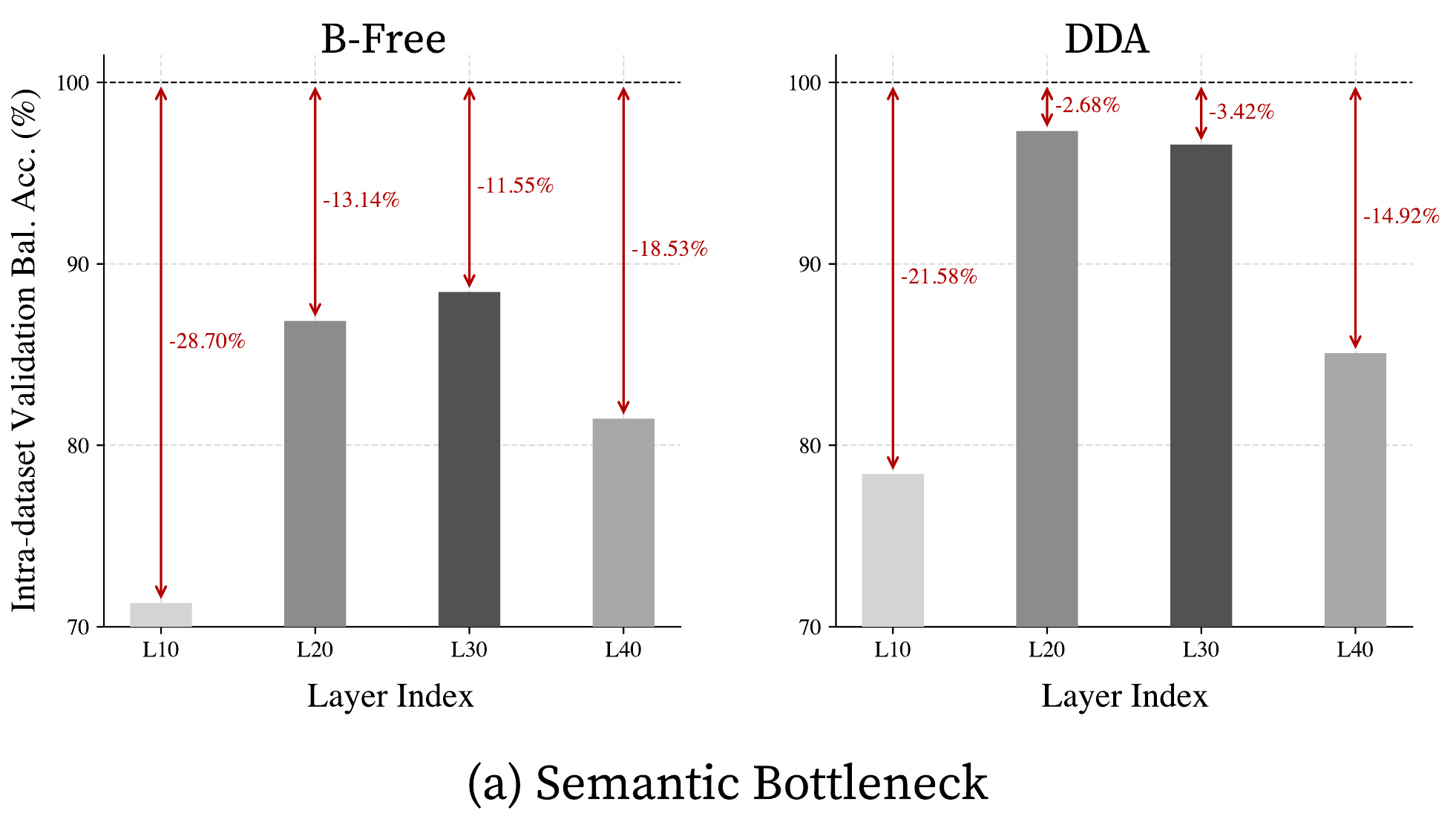}
\caption{Layer-wise probing on B-Free and DDA also reveals a \textit{semantic bottleneck}.}
\Description{Layer-wise probing on B-Free and DDA also reveals a \textit{semantic bottleneck}.}
\label{fig:figure_5}
\end{figure}

\noindent To verify that the observations in Section~\ref{subsec:semantic_bottleneck} of the main paper are not specific to AlignedGenImage, we repeat the same analysis on two additional aligned datasets, B-Free~\cite{guillaro2025bias} and DDA~\cite{chen2025dual}. Figure~\ref{fig:figure_5} reports the probing performance obtained by training independent linear heads on features extracted from different DINOv3 layers. In both datasets, the weakest performance appears at the shallowest layer, improves markedly in the intermediate layers, and then drops again at the final layer. On B-Free, the best probing result is achieved at $L30$, with clear drops at both $L10$ and $L40$. DDA exhibits a similar pattern, with the strongest performance at $L20$--$L30$ and a significant decrease at $L40$. These results provide further evidence that fine-grained pixel-artifact cues can be exploited more effectively in intermediate layers than at the final layer, supporting the semantic bottleneck interpretation.

\begin{figure}[b!]
\centering
\includegraphics[width=1.0\linewidth]{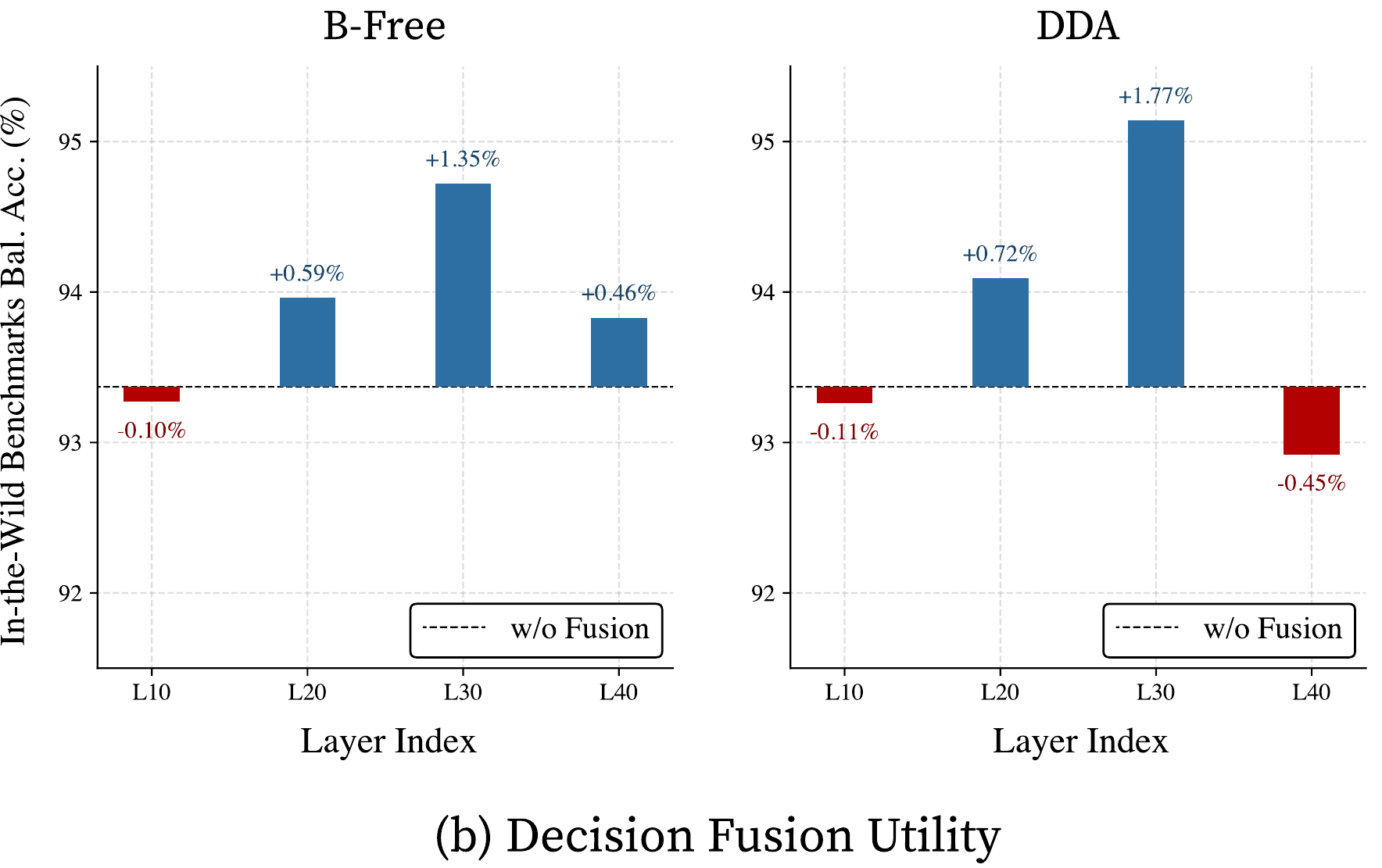}
\caption{Fusing predictions from intermediate experts compensates for this bottleneck, improving generalization.}
\Description{Fusing predictions from intermediate experts compensates for this bottleneck, improving generalization.}
\label{fig:figure_6}
\end{figure}

Figure~\ref{fig:figure_6} further shows that intermediate layer features can be used to improve in-the-wild detection through decision fusion. Specifically, we fuse the predictions of the baseline head trained on standard GenImage final layer features with those of expert heads trained on intermediate layer features from B-Free or DDA. On both datasets, this fusion yields consistent gains over the no-fusion baseline, reaching a maximum value at $L30$ (+1.35\%) on B-Free and $L30$ (+1.77\%) on DDA. In contrast, fusion with the shallowest head ($L10$) slightly degrades performance in both cases, and the gain at $L40$ remains smaller than that of the best intermediate layer. This consistent pattern across all three aligned settings supports the view that fine-grained pixel-artifact cues provide complementary traces for robust in-the-wild detection when combined with the stronger final layer semantics.

\subsection{Detailed Protocol for Geometric Analysis}
\label{sec:geometric_details}
For the geometric analyses in Section~\ref{subsec:motivation}, we use the hybrid composition described in the main paper, consisting of the original real images paired with a 50/50 mixture of standard GenImage fakes and AlignedGenImage fakes. We sample 2,000 test images from each category (real, GenImage fake, and AlignedGenImage fake), and use the same sampled instances across all evaluated settings to ensure a controlled comparison of representation changes. For each case, features are extracted from the final layer of DINOv3-7B with its default pre-trained LayerNorm applied. PCA is performed separately for each case using the real-image features only. Unless otherwise noted, all geometric metrics, including relative drift, intra-class dispersion, and Fisher ratio, are computed in the original feature space after the backbone's final layer normalization, without additional normalization.

\subsection{Analysis of Regularization Variants}
\label{sec:regularization_variants}
In the main paper, we show that naive backbone adaptation can capture subtle forgery cues, but often at the cost of distorting the real-side structure that supports robust generalization. ARA is designed to address this trade-off by combining hybrid data adaptation with an anchor-based regularizer. In this section, we analyze the regularization strategy at two complementary levels. First, we modify the form and support of the regularizer while keeping the backbone adaptation and hybrid training setup unchanged. Second, we compare the full ARA formulation with broader regularized adaptation strategies under the same controlled DINOv3-7B setting. All results are reported using the average performance across the five in-the-wild benchmarks.

\noindent\textbf{Internal Regularization Variants.}
We first compare ARA with three related variants. Real Feature Matching modifies the form of the regularizer, whereas the remaining two variants modify the subset of training samples on which the anchor loss is applied.

\begin{itemize}[leftmargin=1.5em]
\item \textbf{Real Feature Matching.} This variant tests whether the gain of ARA can be explained by generic feature preservation on real samples. It keeps $\mathcal{L}_{\text{main}}$ unchanged and replaces $\mathcal{L}_{\text{anchor}}$ with an $\ell_2$ penalty on real features. Specifically, for each real sample, we penalize the distance between the feature extracted by the adapted backbone and the corresponding feature from the frozen backbone.

\item \textbf{Anchor Loss on Real Only.} This variant tests whether preserving the real-side structure alone is sufficient, without explicitly constraining fake samples. It keeps the total objective in Eq.~\ref{eq:ara_loss} unchanged except for the support of $\mathcal{L}_{\text{anchor}}$. The main classification loss is still computed on all samples, but the anchor loss is evaluated only on the real subset of each mini-batch.

\item \textbf{Anchor Loss on Real and Original Fake.} This variant tests whether extending the anchor loss from real images to standard GenImage fakes is beneficial. It keeps the total objective in Eq.~\ref{eq:ara_loss} unchanged and applies $\mathcal{L}_{\text{anchor}}$ to real images and standard GenImage fakes, while AlignedGenImage fakes are optimized only through $\mathcal{L}_{\text{main}}$.
\end{itemize}

Naive LoRA denotes the $\lambda=0$ setting, where backbone adaptation is performed on the hybrid data without the anchor regularizer.

\begin{table}[h]
\centering
\caption{Comparison of internal regularization variants in balanced accuracy (\%), averaged over the five in-the-wild benchmarks.}
\label{tab:anchor_variants}
\begin{tabular}{p{5.5cm}|c}
\toprule
\textbf{Method} & \textbf{Avg.} \\
\midrule
Naive LoRA ($\lambda=0$) & 92.7 \\
Real Feature Matching & 93.5 \\
Anchor Loss on Real Only & 94.3 \\
Anchor Loss on Real and Original Fake & 94.8 \\
\midrule
\textbf{ARA (Ours)} & \textbf{96.9} \\
\bottomrule
\end{tabular}
\end{table}

As shown in Table~\ref{tab:anchor_variants}, all three variants improve over naive LoRA, but none matches the full ARA formulation. Real Feature Matching reaches 93.5\%, indicating that limiting feature deviation from the frozen backbone is beneficial. Anchor Loss on Real Only reaches 94.3\%, while expanding the support to real and original fake samples increases performance to 94.8\%. The highest performance of 96.9\% is obtained when the anchor loss is applied to the full hybrid training set, including aligned fake samples. These results show that both the form and support of the regularizer affect the resulting generalization performance.

\noindent\textbf{Comparison with Adaptation and Preservation Baselines.}
The preceding experiments examine alternative regularization configurations within the ARA training framework. We next compare ARA with broader adaptation and preservation baselines, including Feature Distillation, Output Distillation, L2-SP, LP-FT, and WiSE-FT. To ensure a controlled comparison, all methods follow the same two-model setup and hybrid-data adaptation protocol as ARA. Specifically, the Stage 1 detector, obtained by training a linear head on the frozen DINOv3-7B backbone using GenImage, is kept fixed and used as a common reference. Separately, a DINOv3-based detector with a newly initialized linear head is adapted with LoRA using the same hybrid dataset as ARA, consisting of real GenImage samples and a fake set composed equally of standard GenImage and AlignedGenImage fakes. Thus, all methods share the same frozen Stage 1 reference and the same adaptation data, while differing in how information from the reference model is preserved or incorporated during adaptation.

For Feature Distillation, the adapted backbone is constrained by an \(\ell_2\) feature-matching loss against the frozen Stage 1 backbone over all samples in the hybrid training set. This differs from Real Feature Matching in the preceding internal ablation, where the main classification objective is likewise optimized on the full hybrid dataset, but the feature-matching constraint is applied only to real samples. Feature Distillation therefore evaluates conventional feature-level preservation under the same full hybrid support used for adaptation, whereas Real Feature Matching specifically examines whether preserving the real-side representation alone is sufficient. The other baselines follow the same controlled setup while replacing ARA's anchor-based regularization with their respective adaptation or preservation strategies.

In addition to balanced accuracy (BACC), we report the area under the receiver operating characteristic curve (AUROC), the area under the precision-recall curve (AUPRC), equal error rate (EER), and the false-positive rate at a 95\% true-positive rate (FPR95). Each metric is averaged over the five in-the-wild benchmarks.

\begin{table}[h]
\centering
\small
\setlength{\abovecaptionskip}{4pt}
\caption{Comparison of regularized adaptation baselines, averaged over the five in-the-wild benchmarks.}
\label{tab:preservation_baselines}
\begin{tabular}{l|ccccc}\toprule
\textbf{Method} & \textbf{BACC$\uparrow$} & \textbf{AUROC$\uparrow$} & \textbf{AUPRC$\uparrow$} & \textbf{EER$\downarrow$} & \textbf{FPR95$\downarrow$} \\
\midrule
Feature Dist. & 89.8 & 98.5 & 98.6 & 5.1 & 8.7 \\
Output Dist. & 92.9 & 98.3 & 98.8 & 4.8 & 12.6 \\
L2-SP & 95.1 & 98.8 & 98.7 & 4.0 & 3.6 \\
LP-FT & 79.2 & 96.0 & 97.1 & 9.3 & 22.0 \\
WiSE-FT & 90.2 & 98.5 & 98.7 & 4.6 & 7.2 \\
\midrule
\textbf{ARA (Ours)} & \textbf{96.9} & \textbf{99.3} & \textbf{99.0} & \textbf{2.7} & \textbf{1.9} \\
\bottomrule
\end{tabular}
\end{table}

As shown in Table~\ref{tab:preservation_baselines}, ARA achieves the best performance across all reported metrics, reaching 96.9\% BACC, 99.3\% AUROC, and 99.0\% AUPRC, with an EER of 2.7\% and an FPR95 of 1.9\%. Among the evaluated baselines, L2-SP provides the strongest balanced accuracy at 95.1\%, but remains 1.8 percentage points below ARA. Compared with L2-SP, ARA also reduces EER and FPR95 by 1.3 and 1.7 percentage points, respectively. Output Distillation achieves the highest AUPRC among the baselines at 98.8\%, but exhibits a substantially higher FPR95 of 12.6\%. The remaining methods show larger gaps in balanced accuracy or false-positive control.

Taken together, these results show that both the design and support of the anchor constraint are important, and that its gains are not reproduced by the evaluated feature-, output-, or parameter-level adaptation strategies. Under the controlled DINOv3-7B setting, ARA provides the best overall balance between detection performance and false-positive control.

\subsection{Effect of Backbone Scale}
\label{sec:backbone_scale}
The main experiments use DINOv3-7B, whose strong frozen
linear-probing performance motivates our analysis of how aligned supervision can be incorporated without degrading the representation structure that supports generalization. We further examine whether the adaptation behavior observed in this setting extends to smaller DINOv3 variants. Specifically, we evaluate DINOv3-H+, DINOv3-L,
and DINOv3-B under the same training and evaluation protocol. Table~\ref{tab:backbone_scales} reports balanced accuracy averaged over the five in-the-wild benchmarks.

\begin{table}[h]
\centering
\caption{Comparison of balanced accuracy (\%) across DINOv3 backbone scales on the in-the-wild benchmark average.}
\label{tab:backbone_scales}
\begin{tabular}{l|ccc}
\toprule
\textbf{Backbone}
& \textbf{Linear Probe}
& \textbf{Naive LoRA}
& \textbf{ARA (Ours)} \\
\midrule
DINOv3-7B & 93.4 & 92.7 & \textbf{96.9} \\
DINOv3-H+ & 87.5 & 84.4 & \textbf{93.2} \\
DINOv3-L & 78.6 & 88.3 & \textbf{91.4} \\
DINOv3-B & \textbf{73.3} & 62.8 & 63.5 \\
\bottomrule
\end{tabular}
\end{table}
\begin{figure*}[t]
\centering
\includegraphics[width=0.7\linewidth]{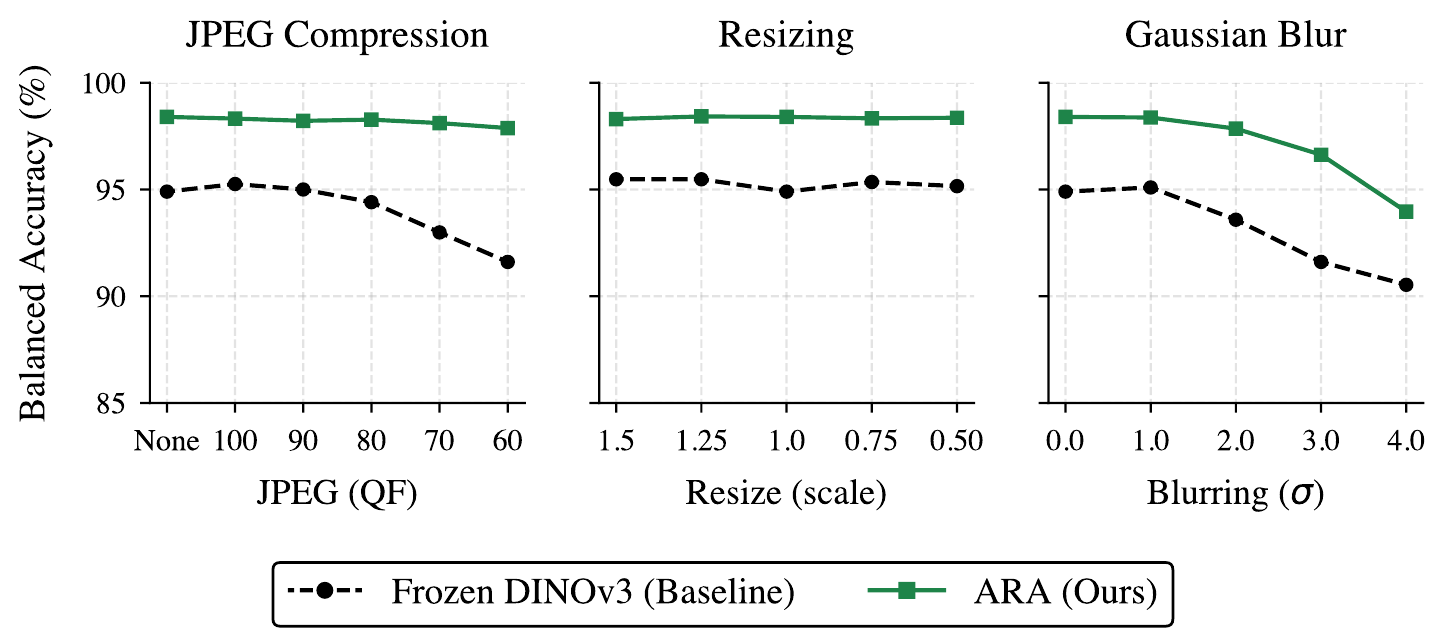}
\caption{Balanced accuracy (\%) under JPEG compression, resizing, and Gaussian blur on Synthbuster. We compare the frozen DINOv3 linear probing baseline with our ARA.}
\Description{Three line plots comparing the frozen DINOv3 linear probing baseline and ARA on Synthbuster under JPEG compression, resizing, and Gaussian blur. The y-axis shows balanced accuracy, and ARA remains consistently above the baseline across perturbation levels.}
\label{fig:figure_7}
\end{figure*}

For each backbone, the linear probe is trained only on GenImage, whereas naive LoRA and ARA use the same hybrid training data. The comparison between naive LoRA and ARA therefore isolates the effect of the anchor regularizer under the same adaptation setting. As shown in Table~\ref{tab:backbone_scales}, ARA consistently outperforms naive LoRA across all evaluated backbone scales. The gains over naive LoRA are 4.2, 8.8, 3.1, and 0.7 percentage points for DINOv3-7B, DINOv3-H+, DINOv3-L, and DINOv3-B, respectively. ARA also outperforms the corresponding frozen linear probe for DINOv3-7B, DINOv3-H+, and DINOv3-L. DINOv3-B exhibits a different pattern: both hybrid adaptation methods perform below the linear probe, although ARA partially mitigates the degradation
observed with naive LoRA. These results show that the anchor regularizer consistently improves over unconstrained hybrid adaptation across the evaluated DINOv3 variants, while the overall benefit of hybrid adaptation becomes limited at the smallest backbone scale.

\subsection{Robustness to Post-Processing}
\label{sec:robustness}
We further evaluate robustness under common post-processing perturbations on Synthbuster~\cite{bammey2023synthbuster}. Since the frozen DINOv3 linear probing detector serves as both the empirical starting point and the analytical baseline of our study, we compare ARA directly against this baseline under JPEG compression, resizing, and Gaussian blur. Following the robustness protocol of B-Free~\cite{guillaro2025bias}, we adopt the same perturbation types and severity settings, and report balanced accuracy (\%) averaged over the Synthbuster generators.

As shown in Figure~\ref{fig:figure_7}, ARA consistently outperforms the frozen DINOv3 baseline across all perturbation types and severity levels. The average balanced accuracy improves from 93.96\% to 97.77\%, with gains of 4.31\% under JPEG compression, 2.99\% under resizing, and 4.00\% under blur. Notably, this improvement is maintained as corruption strength increases, suggesting that the benefits of ARA are not limited to lightly degraded inputs. These results further support our main findings by showing that complementary supervision from misaligned and aligned data can improve detection performance while remaining robust under realistic post-processing conditions.

\end{document}